%% file: acl2018.tex
\documentclass[11pt,a4paper]{article}
\usepackage[hyperref]{acl2018}
\usepackage{times,verbatim}
\usepackage{latexsym}
\usepackage{amsmath}
\usepackage{amssymb}
\usepackage{url}
\usepackage{booktabs}
\usepackage{graphicx}
\usepackage[ruled,linesnumbered]{algorithm2e}
\usepackage{enumitem}
\usepackage{subfigure}
\usepackage{paralist}
\graphicspath{ {images/} }

\aclfinalcopy 

\title{Building Language Models for Text with Named Entities}

\author{
Md Rizwan Parvez \\
University of California Los Angeles \\
 {\tt rizwan@cs.ucla.edu} \\ 
 \\
  {\bf Baishakhi Ray}
\\
Columbia University  \\
 {\tt rayb@cs.columbia.edu} \\ 
 \\ \And Saikat Chakraborty \\
University of Virginia \\
 {\tt saikatc@virginia.edu} \\
 \\
{\bf Kai-Wei Chang}\\
University of California Los Angeles \\
 {\tt kwchang@cs.ucla.edu}
}

\date{}

\begin{document}
\maketitle
\begin{abstract}

Text in many domains involves a significant amount of named entities. Predicting the entity names is often challenging for a language model as they appear less frequent on the training corpus. In this paper, we propose a novel and effective approach to building a discriminative language model which can learn the entity names by leveraging their entity type information. We also introduce two benchmark datasets based on recipes and Java programming codes, on which we evaluate the proposed model. Experimental results show that our model achieves 52.2\% better perplexity in recipe generation and 22.06\% on code generation than the state-of-the-art language models.
\end{abstract}

\section{Introduction}

Language model is a fundamental component in Natural Language Processing (NLP) and it supports various applications, including document generation~\cite{Harvard_data_to_document}, text auto-completion~\cite{ACK17}, spelling correction~\cite{spelling}, and many others. Recently, language models are also successfully used to generate software source code written in programming languages like Java, C, etc.~\cite{naturalnessofsoft,cmu_code_gen,deep_net_for_source_code,code_gen_parsing}. These models have improved the language generation tasks to a great extent, e.g., ~\cite{mikolov2010,Sordoni}. However, while generating text or code with a large number of named entities (e.g., different variable names in source code), these models often fail to predict the entity names properly due to their wide variations.  For instance, consider building a language model for generating recipes. There are numerous similar, yet slightly different cooking ingredients  (e.g., {\it olive oil, canola oil, grape oil, etc.}\textemdash all are different varieties of oil). Such diverse  vocabularies of the ingredient names hinder the language model from predicting them properly. 



To address this problem, we propose a novel language model for texts with many entity names. Our model learns the probability distribution over all the candidate words by leveraging the entity type information. For example, oil is the {\em type} for named entities like olive oil, canola oil, grape oil, etc.\footnote{Entity type information is often referred as category information or group information. In many applications, such information can be easily obtained by an ontology or by a pre-constructed entity table.} 
Such type information is even more prevalent for source code corpus written in statically typed programming languages~\cite{bruce1993safe}, since all the  variables are by construct associated with types like integer, float, string, etc. 

\begin{figure}[t]
\includegraphics[width=7.6cm, height=5.5cm]{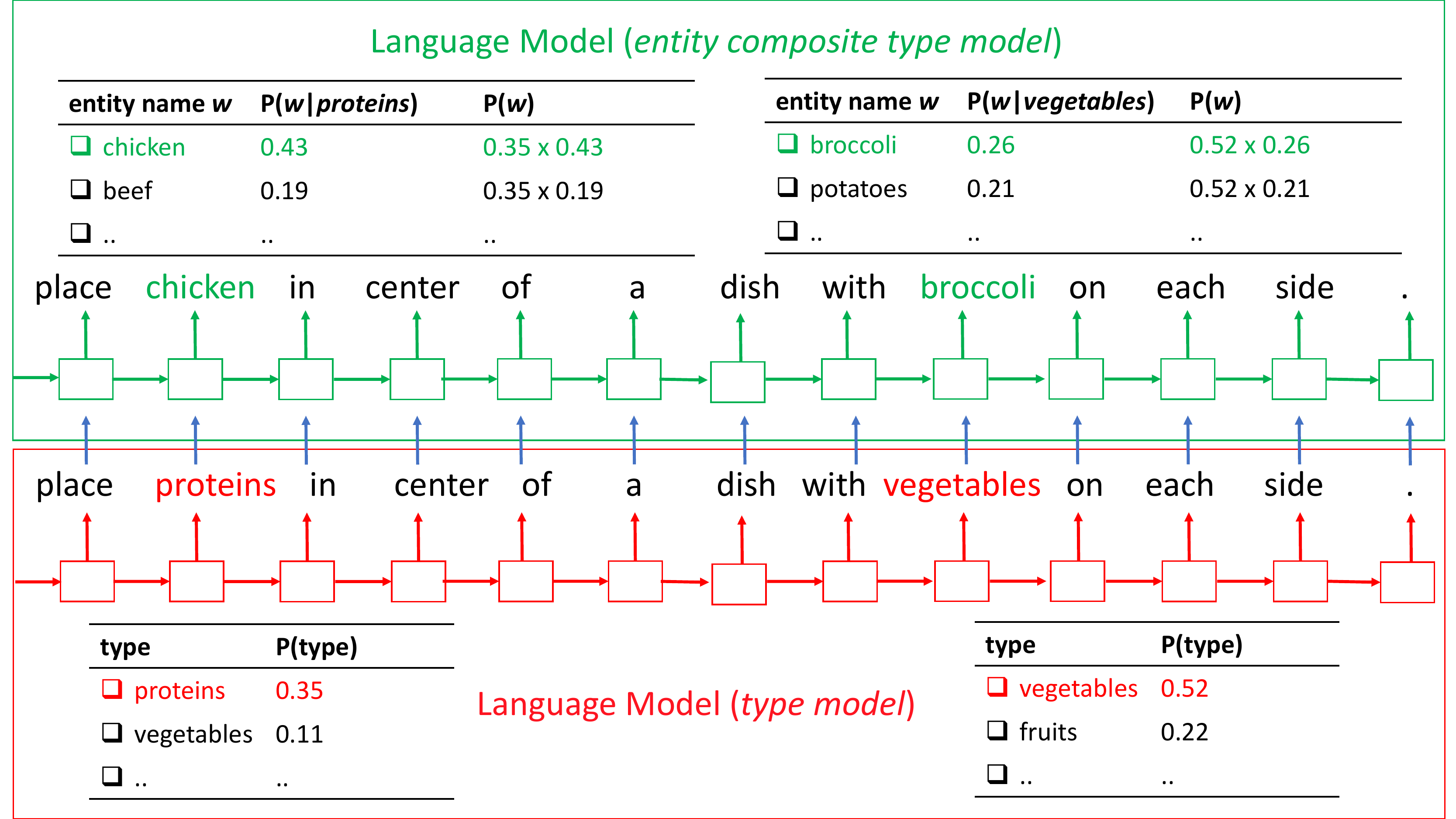}
\caption{\textbf{\small An example illustrates the proposed model. For a given context (i.e., types of context words as input), the \textit{type model} (in  bottom red block)  generates the type of the next word (i.e., the probability of the type of the next word as output). Further, for a given context {\em and} type of each candidate (i.e., context words, corresponding types of the context words, and type of the next word generated by the \textit{type model} as input), the \textit{entity composite model} (in upper green block) predicts the next word (actual entity name) by estimating the conditional probability of the next word as output. The proposed approach conducts joint inference over both models to leverage type information for generating text.}}


  \label{fig:model}
\end{figure}

Our model exploits such deterministic type information of the named entities and learns the probability distribution over the candidate words by decomposing it into two sub-components:
(i) {\em Type Model.}~Instead of distinguishing the individual names of the same type of entities, we first consider all of them equal and represent them by their type information.  This reduces the vocab size to a great extent and enables to predict the type of each entity more accurately.
(ii) {\em  Entity Composite Model.}~Using the entity type as a prior, we learn the conditional probability distribution of the actual entity names at inference time. We depict our model in Fig.~\ref{fig:model}.


To evaluate our model, we create two benchmark datasets that involve many named entities. One is a cooking recipe corpus\footnote{ Data is crawled from \url{http://www.ffts.com/recipes.htm}.} where each recipe contains a number of ingredients which are categorized into 8 super-ingredients (i.e., type);  e.g., ``proteins'', ``vegetables'', ``fruits'', ``seasonings'', ``grains'', etc.  
Our second dataset comprises a source code corpus of  500 open-source Android projects collected from GitHub. We use an Abstract Syntax Tree (AST)~\cite{parsons1992introduction} based approach to collect the type information of the code identifiers. 

Our experiments show that although state-of-the-art language models are, in general, good to learn the frequent words with enough training instances, they perform poorly on the entity names. A simple addition of type information as an extra feature to a neural network does not guarantee to improve the performance because more features may overfit or need more model parameters on the same data. In contrast, our proposed method significantly outperforms state-of-the-art neural network based language models and also the models with type information added as an extra feature. 

Overall, followings are our contributions:

\setlist[itemize]{leftmargin=*}

\begin{compactitem}
    \item We analyze two benchmark language corpora where each consists of a reasonable number of entity names. While we leverage an existing  corpus for recipe, we curated the code corpus. For both datasets, we created auxiliary corpora with entity type information. All the code and datasets are released.\footnote{https://github.com/uclanlp/NamedEntityLanguageModel}

     \item We design a language model for text consisting of many entity names. The model learns to mention entities names by leveraging the entity type information.  
     
     \item We evaluate our model on our benchmark datasets and establish a new baseline performance which significantly outperforms  state-of-the-art language models.
     
\end{compactitem}

\input{related}
\section{A Probabilistic Model for Text with Named Entities}
In this section, we present our approach to build a language model for text with name entities. Given previous context  $\Bar{w} = \{w_1, w_2, .., w_{t-1}\}$, the goal of a language model is to predict the probability of next word $P(w_t | \Bar{w})$ at time step $t$, where $w_t\in V^{text}$ and $V^{text}$ is a fixed vocabulary set. 
Because the size of vocabulary for named entities is large and  named entities often occur less frequently in the training corpus, the language model cannot generate these named entities accurately. For example, in our recipe test corpus the word ``apple'' occurs only 720 times whereas any kind of ``fruits'' occur 27,726 times.  Existing approaches often either only generate common named entities or omit entities when generating text~\cite{jozefowicz2016exploring}. 

To overcome this challenge, we propose to leverage the entity type information when modeling text with many entities. 
We assume each entity is associated with an entity type in a finite set of categories $S = \{s_1,s_2, .., s_i, .., s_k\}$. Given a word $w$, $s(w)$ reflects its entity type. If the word is a named entity, then we denote $s(w) \in S$; otherwise the type function returns the words itself (i.e, $s(w) = w$). To simplify the notations, we use $s(w) \not\in S$ to represent the case where the word is not an entity.
The entity type information given by $s(w)$ is an auxiliary information that we can use to improve the language model. We use $s(\bar{w})$ to represent the entity type information of all the words in context $\bar{w}$ and use $w$ to represent the current word $w_t$. Below, we show that a language model for text with typed information can be decomposed into the following two models: 1) a {\it type model $\theta_t$} that predicts the entity type of the next word and 2) an {\it entity composite model $\theta_v$} that predicts the next word based on a given entity type.  

Our goal is to model the probability of next word $w$ given previous context $\bar{w}$:
\begin{equation}
\label{eq:goal}
P\left(w| \Bar{w}; \theta_t, \theta_v \right),
\end{equation}
where $\theta_t$ and $\theta_v$ are the parameters of the two aforementioned models.
As we assume the typed information is given on the data,
Eq. \eqref{eq:goal} is equivalent to  
\begin{equation}
\label{eq: goal2}
 P\left(w, s(w) | \Bar{w}, s(\Bar{w}); \theta_t, \theta_v \right).
\end{equation}

A word can be either a named entity or not; therefore, we consider the following two cases.  

\paragraph{\bf Case 1: next word is a named entity.} In this case, Eq. \eqref{eq: goal2} can be rewritten as 
\begin{equation}
\label{eq:entity}
    \begin{split}
    &P\left(s(w)=s |\Bar{w}, s(\Bar{w}); \theta_t, \theta_v \right) \times\\
    &P\left(w|\Bar{w}, s(\Bar{w}), s(w)=s; \theta_v, \theta_t \right)          
    \end{split}
\end{equation}
based on the rules of conditional probability. 

We assume the type of the next token $s(w)$ can be predicted by a model $\theta_t$ using information of $s(\bar{w})$, and we can approximate the first term in Eq. \eqref{eq:entity} 
\begin{equation}\label{eq:approx_assumption}
    P(s(w)|\Bar{w},s(\Bar{w}); \theta_t, \theta_v) \approx P(s(w)|s(\Bar{w}), \theta_t)
\end{equation}
Similarly, we can make a modeling assumption to simplify the second term as 
\begin{equation}\label{eq:var_approx_assumption}
\begin{split}
P(w|\Bar{w},s(\Bar{w}), s(w), \theta_v, \theta_t) \\ \approx  P(w|\Bar{w}, s(\bar{w}), s(w), \theta_v).
\end{split}
\end{equation}

\paragraph{\bf Case 2: next word is not a named entity.}  In this case, we can rewrite Eq. \eqref{eq: goal2} to be
\begin{equation}
\label{eq:common}
    \begin{split}
  &  P\left(s(w) \not \in S|\Bar{w},  s(\bar{w}), \theta_t \right) \times\\
 & P\left(w|\Bar{w}, s(\Bar{w}), s(w) \not \in S, \theta_v \right).
 \end{split}
\end{equation}

The first term in Eq. \eqref{eq:common} can be modeled by 
\begin{equation*}
1- \sum_{s \in S} P(s(w) = s | s(\Bar{w}), \theta_t),
\end{equation*}
which can be computed by the {\it type} model\footnote{Empirically for the non-entity words, $\sum_{s \in S} P(s(w) = s | s(\Bar{w}) \approx 0$}. 
The second term can be again approximated by \eqref{eq:var_approx_assumption} and further estimated by an {\it entity composition model}.

\paragraph{\bf Typed Language Model.} Combine the aforementioned equations, the proposed language model estimates $P(w| \bar{w}; \theta_t, \theta_v)$ by 
\begin{equation}
\label{eq:combine}
\begin{split}
&P(w|\Bar{w}, s(\bar{w}), s(w), \theta_v) \times\\
&  \begin{cases}
 P(s(w)|s(\Bar{w}), \theta_t) 
 \  &\text{if $s(w)  \in S$ } \\
 (1\!-\!\sum_{s \in S} P(s(w)\!=\! s | s(\Bar{w}), \theta_t)) \  &
 \text{if $s(w)  \not \in S$ } \\
  \end{cases}  
  \end{split}
\end{equation}

The first term can be estimated by an {\it entity composite model}  and the second term can be estimated by a {\it type model} as discussed below.

\subsection{ Type model}
\label{type model}

The \textit{ type model} $\theta_t$ estimates the probability of $P(s(w)|s(\bar{w}), \theta_t)$. It can be viewed as a language model builds on a corpus with all entities replaced by their type. That is, assume the training corpus consists of $x = \{ w_1, w_2, .., w_n \}$. Using the type information provided in the auxiliary source, we can replace each word $w$ with their corresponding type $s(w)$ and generate a corpus of $\mathcal{T} = \{ s(w_i), s(w_2), .., s(w_n)\}$. Note that if $w_i$ is not an named entity (i.e., $s(w) \not\in S$), $s(w) = w$ and the vocabulary  on $\mathcal{T}$ is $V^{text} \cup S$.\footnote{In a preliminary experiment, we consider putting all words with $s(w) \not\in S$ in a category ``N/A''. However, because most words on the training corpus are not named entities, the type ``N/A'' dominates others and hinder the \textit{ type model} to make accurate predictions. } Any language modeling technique can be used in modeling the \textit{ type model} on the modified corpus $\mathcal{T}$. In this paper, we use the state-of-the-art model for each individual task. The details will be discussed in the experiment section. 






\subsection{Entity Composite Model}
\label{entity composite model}

The {\it entity composite model} predicts the next word based on modeling the conditional probability 
$P(w|\Bar{w}, s(\bar{w}), s(w), \theta_v)$, which can be derived by
\begin{equation}
\label{eq:7}
\frac{P(w|\Bar{w}, s(\bar{w}); \theta_v)}
{\sum_{w_s\in \Omega(s(w))} P(w_s|\Bar{w}, s(\bar{w}); \theta_v)},
\end{equation}
where $\Omega(s(w))$ is the set of words of the same type with $w$.

To model the types of context word $s(\bar{w})$ in $P(w|\Bar{w}, s(\bar{w}); \theta_v)$, we consider learning a type embedding along with the word embedding by augmenting each word vector with a type vector when learning the underlying word representation. Specifically, we represent each word $w$ as a vector of $[v_w(w)^T; v_t(s(w))^T]^T$, where $v_w(\cdot)$ and $v_t(\cdot)$ are the word vectors and type vectors learned by the model from the training corpus, respectively. 
Finally, to estimate Eq. \eqref{eq:7} using $\theta_v$, when computing the Softmax layer, we normalize over only words in $\Omega(s(w))$. In this way, the conditional probability 
$P(w|\Bar{w}, s(\bar{w}), s(w), \theta_v)$ can be derived. 





\subsection{Training and Inference Strategies}

We learn model parameters $\theta_t$ and $\theta_v$ independently by training two language models {\it type model} and {\it entity composite model} respectively. Given the context of type, {\it type model} predicts the type of the next word. Given the context and the type information of the all candidate words, {\it entity composite model} predicts the conditional actual word (e.g., entity name) as depicted in Fig \ref{fig:model}. At inference time the generated probabilities from these two models are combined according to conditional probability (i.e., Eq. (\ref{eq:combine})) which gives the final probability distribution over all candidate words\footnote{While calculating the final probability distribution over all candidate words, with our joint inference schema, a strong state-of-art language model, without the type information, itself can work sufficiently well and replace the {\it entity composite model}. Our experiments using \cite{socher} in Section \ref{recipe_gen} validate this claim.}. 

Our proposed model is flexible to any language model, training strategy, and optimization. As per our experiments, we use ADAM stochastic mini-batch optimization~\cite{adam}. 
In Algorithm \ref{algo-1}, we summarize the language generation procedure.  


\begin{algorithm}[t]
\DontPrintSemicolon
\KwIn{Language corpus $\mathcal{X} = \{w_1,w_2,..,w_n\}$, 
   type $s(w)$ of the words, integer number $m$.}
\KwOut{$\theta_t$, $\theta_v$, \{$W_1,W_2,.., W_m$\}}
\vspace{5pt}
 {\bf Training Phase: \;}{
 Generate $\mathcal{T}$ = \{ $s(w_1), s(w_2), .., s(w_n)$\}\;
 Train \textbf{type model} $\theta_t$ on $\mathcal{T}$\;
 Train \textbf{entity composite model} $\theta_v$ on $\mathcal{X}$ using  [$w_i; s(w_i)$] as input\;
 }
 \vspace{5pt}
 {\bf Test Phase (Generation Phase):\;} {
 \For{$i=1$ to $m$} {
 \For{$w \in V^{text}$}{
    Compute $P(s(w)|s(\bar{w}), \theta_t)$\;
    Compute $P(w|\bar{w},s(\bar{w}), s(w), \theta_v)$\;
    Compute $P(w|\bar{w}; \theta_t, \theta_v)$ using Eq.\eqref{eq:combine} \;
    }
  $W_i \gets {argmax}_w P(w| \bar{w}; \theta_t, \theta_v)$ \;
  }
 }
 \caption{\textbf{\small Language Generation}}
 \label{algo-1}
\end{algorithm}



\section{Experiments}
\label{experiments}

We evaluate our proposed model on two different language generation tasks where there exist a lot of entity names in the text. In this paper, we release all the codes and datasets.
The first task is recipe generation. For this task, we analyze a cooking recipe corpus. Each instance in this corpus is an individual recipe and 
consists of 
many ingredients'.
Our second task is code generation. We construct a Java code corpus where each instance is a Java method (i.e., function). 
These tasks are challenging because they have the abundance of entity names and state-of-the-art language models fail to predict them properly as a result of insufficient training observations.  Although in this paper, we manually annotate the types of the recipe ingredients, in other applications it can be acquired automatically.  For example: in our second task of code generation, the types are found using Eclipse JDT framework. In general, using DBpedia ontology (e.g., ``Berlin'' has an ontology ``Location''), Wordnet hierarchy (e.g., ``Dog'' is an ``Animal''), role in sports (e.g., ``Messi'' plays in ``Forward''; also available in DBpedia\footnote{ http://dbpedia.org/page/Lionel\_Messi}), Thesaurus (e.g., ``renal cortex'', ``renal pelvis'', ``renal vein'', all are related to ``kidney''), Medscape (e.g., ``Advil'' and ``Motrin'' are actually ``Ibuprofen''), we can get the necessary type information. As for the applications where the entity types cannot be extracted automatically by these frameworks (e.g., recipe ingredients), although there is no exact strategy, any reasonable design can work. Heuristically, while annotating manually in our first task, we choose the total number of types in such a way that each type has somewhat balanced (similar) size. 

We use the same dimensional word embedding (400 for recipe corpus, 300 for code corpus) to represent both of the entity name (e.g., ``apple'') and their entity type (e.g., ``fruits'') in all the models. Note that in our approach, the \textit{type model} only replaces named entities with entity type when it generates next word. If next word is not a named entity, it will behave like a regular language model. Therefore, we set both models with the same dimensionality. Accordingly, for the \textit{entity composite model} which takes the concatenation of the entity name and the entity type, the concatenated input dimension is 800 and 600 respectively for recipe and code corpora.

\subsection{Recipe Generation}
\label{recipe_gen}
\textbf{ Recipe Corpus Pre-processing:} Our recipe corpus collection is inspired by \cite{recipe_checklist}. We crawl the recipes from ``Now You’re Cooking! Recipe Software'' \footnote{http://www.ffts.com/recipes.htm}. Among more than 150,000 recipes  in this dataset, we select similarly  structured/formatted (e.g, title, blank line then ingredient lists followed by a recipe) 95,786 recipes. We remove all the irrelevant  information (e.g., author's name, data source) and keep only two information: ingredients and recipes.  We set aside the randomly selected 20\% of the recipes for testing and from the rest, we keep randomly selected 80\% for the training and 20\% for the development.  
Similar to \cite{recipe_checklist},  we pre-process the  dataset and  filter out the numerical values, special tokens, punctuation, and symbols.\footnote{For example, in our crawled raw dataset, we find that some recipes have lines like ``===MMMMM==='' which are totally irrelevant to our task. For the words with numerical values like ``100 ml'', we only remove the ``100'' and keep the ``ml'' since our focus is not to predict the exact number.}  Quantitatively, the data we filter out is negligible; in terms of words, we keep 9,994,365 words out of 10,231,106 and the number of filter out words is around $\sim$2\%. We release both of the raw and cleaned data for future challenges. 
As the ingredients are the entity names in our dataset, we process it separately to get the type information.

\textbf{Retrieving Ingredient Type:} As per our type model, for each word $w$, we require its type $s(w)$. We only consider ingredient type for our experiment. First, we tokenize the ingredients and consider each word as an ingredient. We manually classify the ingredients into 8 super-ingredients: ``fruits'', ``proteins'', ``sides'', ``seasonings'', ``vegetables'', ``dairy'', ``drinks'', and ``grains''. Sometimes, ingredients are expressed using multiple words; for such ingredient phrase, we classify each word in the same group (e.g., for ``boneless beef'' both ``boneless'' and ``beef'' are classified as ``proteins''). We classify the most frequent 1,224 unique ingredients, \footnote{We consider both singular and plural forms. The number of singular formed annotated ingredients are 797.}  which cover 944,753  out of  1,241,195 mentions (top 76\%) in terms of frequency of the ingredients. In our experiments, we omit the remainder 14,881 unique ingredients which are less frequent and include some misspelled words. The number of unique ingredients in the 8 super ingredients is 110, 316, 140, 180, 156, 80, 84, and  158 respectively.  We prepare the modified {\it type corpus} by replacing each actual ingredient's name $w$ in the original recipe corpus by the type (i.e., super ingredients  $s(w)$) to train the {\it type model}.


\textbf{Recipe Statistics:} In our corpus, the total number of distinct words in vocabulary is 52,468; number of unique ingredients (considering splitting  phrasal ingredients also) is 16,105; number of tokens is  8,716,664. In number of instances train/dev/test splits are  61,302/15,326/19,158. The average instance size of a meaningful recipe is 91 on the corpus. 

\textbf{Configuration:} \label{config}
We consider the state-of-the art LSTM-based language model proposed in \cite{socher} as the basic component for building the {\it type model}, and {\it entity composite model}.
We use 400 dimensional word embedding as described in Section \ref{experiments}. We train the embedding for our dataset. We use a minibatch of 20 instances while training and  back-propagation through time value is set to 70. Inside of this \cite{socher} language model, it uses 3 layered LSTM architecture where the hidden layers are 1150 dimensional and has its own optimization and regularization mechanism. All the experiments are done using PyTorch and Python 3.5.

\textbf{Baselines:} Our first baseline is ASGD Weight-Dropped LSTM (AWD\_LSTM)~\cite{socher}, which we also use to train our models (see 'Configuration' in \ref{config}). This model achieves the state-of-the-art performance on benchmark Penn Treebank (PTB), and WikiText-2 (WT2) language corpus. 
Our second baseline is the same language model (AWD\_LSTM) with the type information added as an additional feature (i.e., same as {\it entity composite model}). 


\textbf{Results of Recipe Generation.} We compare our model with the baselines using {\em perplexity} metric\textemdash lower perplexity means the better prediction.  Table \ref{rcp_ppl_table} summarizes the result. The 3$^{rd}$ row shows that adding type as a simple feature does not guarantee  a significant performance improvement while our proposed method significantly outperforms both baselines and achieves 52.2\% improvement with respect to  baseline in terms of perplexity. To illustrate more, we provide an example snippet of our test corpus:
``place onion and ginger inside chicken  . allow chicken to marinate for hour  .''. Here, for the last mention of the word ``chicken'', the standard language model assigns probability 0.23 to this word, while ours assigns probability 0.81.

\begin{table}
\scriptsize
\centering

 \begin{tabular}{l| l| c |  c} 
 \toprule
 Model & Dataset  & Vocabulary & Perplexity   \\ 
       & (Recipe Corpus)  & Size &    \\ 
 \midrule
  AWD\_LSTM & original  & 52,472  & 20.23 \\
   \midrule
  AWD\_LSTM  & modified type  &  51,675 & 17.62 \\
  \textit{type model} & & & \\
   \midrule
AWD\_LSTM  & original  &  52,472 & 18.23\\
   with type feature & & & \\
    \midrule
 our model  & original  &  52,472 &  \textbf{9.67} \\
 \bottomrule
 \end{tabular}
 \caption{\textbf{\small Comparing the performance of recipe generation task. All the results are on the test set of the corresponding corpus. AWD\_LSTM (\textit{type model}) is our \textit{type model} implemented with the baseline language model AWD\_LSTM~\cite{socher}. Our second baseline is the same language model (AWD\_LSTM) with the type information added as an additional feature for each word.}}
 
 \label{rcp_ppl_table}
\end{table}
\subsection{Code Generation}
\label{code}

\textbf{Code Corpus Pre-processing.} We crawl 500 Android open source projects from GitHub\footnote{https://github.com}. 
GitHub is the largest open source software forge where anyone can contribute~\cite{ray2014large}. Thus, GitHub also contains trivial projects like student projects, etc. In our case, we want to study the coding practices of practitioners so that our model can learn to generate quality code. To ensure this, we choose only those Android projects from GitHub that are also present in Google Play Store\footnote{https://play.google.com/store?hl=en}. We download the source code of these projects from GitHub using an off the shelf tool GitcProc~\cite{casalnuovo2017gitcproc}.

Since real software continuously evolves to cater new requirements or bug fixes, to make our modeling task more realistic, we further study different project versions. We partition the codebase of a project into multiple versions based on the code commit history retrieved from GitHub; each version is taken at an interval of 6 months. For example, anything committed within the first six months of a project will be in the first version, and so on. We then build our code suggestion task mimicking how a developer develops code in an evolving software\textemdash based on the past project history, developers add new code. To implement that we train our language model on past project versions and test it on the most recent version, at method granularity. However, it is quite difficult for any language model to generate a method from the scratch if the method is so new that even the method signature (i.e., method declaration statement consisting of method name and parameters) is not known. Thus, during testing, we only focus on the methods that the model has seen before but some new tokens are added to it. This is similar to the task when a developer edits a method to implement a new feature or bug-fix.

Since we focus on generating the code for every method, we train/test the code prediction task at method level\textemdash each method is similar to a sentence and each token in the method is equivalent to a word. Thus, we ignore the code outside the method scope like global variables, class declarations, etc. We further clean our dataset by removing user-defined ``String'' tokens as they increase the diversity of the vocabularies significantly, although having the same type.  For example, the word sequences ``Hello World!'' and  ``Good wishes for ACL2018!!'' have the same type {\tt java.lang.String.VAR}. 


\textbf{Retrieving Token Type:} 
For every token $w$ in a method, we extract its type information $s(w)$. A token type can be Java built-in data types (e.g., \textit{int, double, float, boolean} etc.,) or user or framework defined classes (e.g., {\it java.lang.String, io.segment.android.flush.FlushThread} etc.). We extract such type information for each token by parsing the Abstract Syntax Tree (AST) of the source code\footnote{AST represents source code as a tree by capturing its abstract syntactic structure, where each node represents a construct in the source code.}. We extract the  AST type information of each token using Eclipse JDT framework\footnote{https://www.eclipse.org/jdt/}. Note that, language keywords like {\tt for}, {\tt if}, etc. are not associated with any type. 
Next, we prepare the type corpus by replacing the variable names with corresponding type information. For instance, if variable {\tt var} is of type {\it java.lang.Integer}, in the type corpus we replace {\it var} by {\it java.lang.Integer}. Since multiple packages might contain classes of the same name, we retain the fully qualified name for each type\footnote{Also the AST type of a very same variable may differ in two different methods. Hence, the context is limited to each method.}.

\textbf{Code Corpus Statistics:} \label{stat_code} 
 In our corpus, the total number of distinct words in vocabulary is 38,297; the number of unique AST type (including all user-defined classes) is 14,177; the number of tokens is  1,440,993. The number of instances used for train and testing is 26,600 and 3,546. Among these 38,297 vocabulary words, 37,411 are seen at training time while the rests are new.

\textbf{Configuration:} \label{config2}
To train both {\it type} {\it model} and {\it entity composite model}, we use forward and backward LSTM (See Section \ref{related}) and combine them at the inference/generation time. We train 300-dimensional word embedding for each token as described in Section \ref{experiments} initialized by GLOVE~\cite{glove}. Our LSTM is single layered and the hidden size is 300. We implement our model on using PyTorch and Python 3.5. 
Our training corpus size 26,600 and we do not split it further into smaller train and development set; rather we use them all to train for one single epoch and record the result on the test set.


\textbf{Baselines:} 
Our  first baseline is standard LSTM language model which we also use to train our modules (see `Configuration' in \ref{config2}).  Similar to our second baseline for recipe generation we also consider LSTM with the type information added as more features\footnote{LSTM with type is same as {\it entity composite  model}.} as our another baseline. We further compare our model with state-of-the-art token-based language model for source code SLP-Core~\cite{deep_net_for_source_code}.

\begin{table}
\scriptsize
\centering
 \begin{tabular}{l| l| c |  c} 
 \toprule
 Model & Dataset  & Vocabulary & Perplexity   \\ 
       & (Code Corpus)  & Size &    \\ 
       \midrule
       SLP-Core & original   & 38,297  & 3.40 \\
 \midrule
   fLSTM & original   & 38,297  & 21.97 \\
    fLSTM [\textit{type model}] & modified  type  & 14,177  & 7.94 \\
    fLSTM with type feature & original   & 38,297  & 20.05 \\
     our model (fLSTM)  & original  &  38,297 & \textbf{12.52} \\
   \midrule
 bLSTM & original  &  38,297 & 7.19 \\
 
 bLSTM [\textit{type model}] & modified  type  &  14,177 & 2.58 \\
 
 bLSTM with type feature & original  &  38,297 & 6.11 \\
 
  our model (bLSTM)  & original  &  38,297 & \textbf{2.65}  \\
 \bottomrule
 \end{tabular}
 \caption{\textbf{\small Comparing the performance of code generation task. All the results are on the test set of the corresponding corpus. fLSTM, bLSTM denotes forward and backward LSTM respectively. SLP-Core refers to \cite{deep_net_for_source_code}.}}
 \label{code_ppl_table}
\end{table} 

\textbf{Results of Code Generation:}
 Table \ref{code_ppl_table} shows that adding type as simple features does not guarantee a significant performance improvement while our proposed method significantly outperforms both forward and backward LSTM baselines. Our approach with backward LSTM has 40.3\% better perplexity than original backward LSTM and forward has  63.14\% lower (i.e., better) perplexity than original forward LSTM. With respect to SLP-Core performance, our model is 22.06\% better in perplexity. We compare our model with SLP-Core details in case study-2.


\section{Quantitative Error Analysis}
To understand the generation performance of our model and interpret the meaning of the numbers in Table~\ref{rcp_ppl_table} and~\ref{code_ppl_table}, we further perform the following case studies. 

\subsection{Case Study-1: Recipe Generation}

As the reduction of the perplexity does not necessarily mean the improvement of the accuracy, we design a ``fill in the blank task'' task to evaluate our model. 
A blank place in this task will contain an ingredient and we check whether our model can predict it correctly. In particular, we choose six ingredients from different frequency range (low, mid, high) based on how many times they have appeared in the training corpus. Following 
Table shows two examples with four blanks (underlined with the true answer).

\begin{center} 
\begin{tabular}{ l } 
\toprule
 \multicolumn{1}{c}{\textbf{\small Example fill in the blank task} } \\
 \midrule
 1.~Sprinkle \underline{chicken} pieces  lightly with \underline{salt}. \\  
 2.~Mix egg and \underline{milk} and pour over \underline{bread}.  \\
\bottomrule
\end{tabular}
\end{center}
\vspace{2pt}

We further evaluate our model on a multiple choice questioning (MCQ) strategy where the fill in the blank problem remains same but the options for the correct answers are restricted to the six ingredients. 
Our intuition behind this case-study is to check when there is an ingredient whether our model can learn it. If yes, we then quantify the learning using standard {\em accuracy} metric and compare with the state-of-the-art model to evaluate how much it improves the performance. We also measure how much the accuracy improvement depends on the training frequency.


\setlength{\tabcolsep}{2pt}
\begin{table}[h]
\scriptsize
\centering
 \begin{tabular}{l| l| l|l  c|l c} 
 \toprule
   &  &   & \multicolumn{4}{c}{Accuracy}\\
 Ingredient & Train Freq. & \#Blanks  & \multicolumn{2}{c}{Free-Form} & \multicolumn{2}{|c}{MCQ}\\
 &  & & AWD\_LSTM & Our  & AWD\_LSTM & Our\\ 
 \midrule
  Milk  &  14, 136 &4,001&    26.94   &  \textbf{59.34} &  80.83  & \textbf{94.90 }\\
    Salt &  33,906& 9,888&   37.12 & \textbf{62.47 }&89.29 &\textbf{95.75} \\
 Apple & 7,205 & 720  & 1.94    &  \textbf{30.28} &  37.65 &\textbf{89.86} \\
 Bread &11,673 & 3,074 &  32.43  &  \textbf{52.64 }& 78.85 &\textbf{94.53 }\\\
 Tomato &12,866& 1,815&  2.20    &\textbf{35.76 }& 43.53  &\textbf{88.76} \\
 Chicken & 19,875& 6,072 & 22.50  &\textbf{45.24 } &  77.70  &\textbf{94.63} \\
 \bottomrule
 \end{tabular}
 \caption{\textbf{\small Performance of fill in the blank task.}}
 \label{table:case-rcp}
\end{table}

Table \ref{table:case-rcp} shows the result. Our model outperforms the fill in the blank task for both cases, i.e.,  without any options (free-form) and MCQ. Note that, the percentage of improvement is inversely proportional to the training frequencies of the ingredients\textemdash less-frequent ingredients achieve a higher accuracy improvement (e.g., ``Apple'' and ``Tomato''). This validates our intuition of learning to predict the type first more accurately with lower vocabulary set and then use conditional probability to predict the actual entity considering the type as a prior. 

\subsection{Case Study-2: Code Generation} 

Programming language source code shows regularities both in local and global context (e.g., variables or methods used in one source file can also be created or referenced from another library file). SLP-Core~\cite{deep_net_for_source_code} is a state-of-the-art code generation model that captures this global and local information using a nested cache based n-gram language model. They further show that considering such code structure into account, a simple n-gram based SLP-Core outperforms vanilla deep learning based models like RNN, LSTM, etc. 

In our case, as our example instance is a Java method, we only have the local context. Therefore, to evaluate the efficiency of our proposed model, we further analyze that exploiting only the type information are we even learning any global code pattern? If yes, then how much in comparison to the baseline (SLP-Core)?
To investigate these questions, we provide all the full project information to SLP-Core \cite{deep_net_for_source_code} corresponding to our train set. However, at test-time, to establish a fair comparison, we consider the perplexity metric for the same methods. SLP-Core achieves a perplexity 3.40  where our backward LSTM achieves 2.65. This result shows that appropriate type information can actually capture many inherent attributes which can be exploited to build a good language model for programming language.

\section{Conclusion}
\label{ssec:conclusion}
Language model often lacks in performance to predict entity names correctly. Applications with lots of named entities, thus, obviously suffer. In this work, we propose to leverage the type information of such named entities to build an effective language model. Since similar entities have the same type, the vocabulary size of a type based language model reduces significantly. The prediction accuracy of the type model increases significantly with such reduced vocabulary size. Then, using the entity type information as prior we build another language model which predicts the true entity name according to the conditional probability distribution. Our evaluation and case studies confirm that the type information of the named entities captures inherent text features too which leads to learn intrinsic text pattern and improve the performance of overall language model.

\section*{Acknowledgments}
We thank the anonymous reviewers for their insightful comments. We also thank Wasi Uddin Ahmad, Peter Kim, Shou-De Lin, and Paul Mineiro for helping us implement, annotate, and design the experiments. This work was supported in part by National Science Foundation Grants IIS-1760523, CCF-16-19123,  CNS-16-18771 and an NVIDIA hardware grant.

\bibliography{acl2018}
\bibliographystyle{acl_natbib}




\end{document}

%% file: related.tex
\section{Related Work and Background}
\label{related}

\paragraph{\bf Class Based Language Models.}
Building language models by leveraging the deterministic or probabilistic class properties of the words (a.k.a, class-based language models) is an old idea~\cite{brown1992, goodman}. However, the objective of our model is different from the existing class-based language models. 
The key differences are two-folds:
1) 
Most existing class-based language models~\cite{brown1992, pereira1993, niesler1998, baker1998, goodman, maltese2001} are generative n-gram models whereas ours is a discriminative language model based on neural networks. The modeling principle and assumptions are very different. For example, we cannot calculate the conditional probability by statistical occurrence counting as these papers did. 2) Our approaches consider building two models and perform joint inference which makes our framework general and easy to extend. In Section \ref{experiments}, we demonstrate that our model can be easily incorporated with the state-of-art language model.  
The closest work in this line is hierarchical neural language models~\cite{hierarchical}, which model language with word clusters. However, their approaches do not focus on dealing with named entities as our model does. 
A recent work~\cite{dynamic_entity} studied the problem of building up a dynamic representation of named entity by updating the representation for every contextualized mention of that entity.  Nonetheless, their approach does not deal with the sparsity issue and their goal is different from ours.


\paragraph{\bf Language Models for Named Entities.} 
In some generation tasks, recently developed language models address the problem of predicting entity names  by copying/matching the entity  names  from  the  reference  corpus. For example, ~\newcite{pointer_network}  calculates the conditional probability of discrete output token sequence  corresponding to positions in an input sequence. ~\newcite{copy_seq-to_seq} develops a seq2seq alignment mechanism which directly copies entity names or long phrases from the input sequence. \newcite{Harvard_data_to_document} generates document from  structured table like basketball statistics using copy and reconstruction method as well. Another related code generation model~\cite{cmu_code_gen}  parses natural language descriptions into source code considering the grammar and syntax in the target programming language  (e.g., Python). ~\newcite{recipe_checklist} generates recipe for a given goal, and agenda by making use of items on the agenda. While generating the recipe it continuously monitors the agenda coverage and focus on increasing it. All of them are sequence-to-sequence learning or end-to-end systems which differ from our general purpose  (free form) language generation task (e.g., text auto-completion, spelling correction). 


\textbf{Code Generation.} The way developers write codes is not only just writing a bunch of instructions to run a machine, but also a form of communication to convey their thought. As observed by Donald E.~Knuth~\cite{knuth1992literate}, ~\textit{``The practitioner of literate programming can be regarded as an essayist, whose main concern is exposition and excellence of style. Such an author, with thesaurus in hand, chooses the names of variables carefully and explains what such variable means."} Such comprehensible software corpora show surprising regularity ~\cite{ray2014uniqueness,gabel2010study} that is quite similar to the statistical properties of natural language corpora and thus, amenable to large-scale statistical analysis~\cite{hindle2012naturalness}. ~\cite{allamanis2017survey} presented a detailed survey.

Although similar, source code has some unique properties that differentiate it from natural language. For example, source code often shows more regularities in local context due to common development practices like copy-pasting~\cite{gharehyazie2017some,kim2005empirical}. This property is successfully captured by cache based language models~\cite{deep_net_for_source_code,tu2014localness}. 
Code is also less ambiguous than natural language so that it can be interpreted by a compiler. The constraints for generating correct code is implemented by combining language model and program analysis technique~\cite{raychev2014code}. Moreover, code contains open vocabulary\textemdash developers can coin new variable names without changing the semantics of the programs. Our model aims to addresses this property by leveraging variable types and scope.

\paragraph{\bf LSTM Language Model.}
In this paper, we use LSTM language model as a running example to describe our approach.
Our language model uses the LSTM cells to generate latent states for a given context which captures the necessary features from the text. At the output layer of our model, we use Softmax probability distribution to predict the next word based on the latent state. ~\newcite{socher} is a LSTM-based language model which achieves the state-of-the-art performance on Penn Treebank (PTB) and WikiText-2 (WT2) datasets. To build our recipe language model we use this as a blackbox and for our code generation task we use the simple LSTM model both in forward and backward direction. A forward directional LSTM starts from the beginning of a sentence and goes from left to right sequentially until the sentence ends, and vice versa. However, our approach is general and can be applied with other types of language models. 